\documentclass[journal,twoside]{IEEEtran}
\ifCLASSINFOpdf
\else
\fi
\usepackage{amssymb}
\usepackage{amsmath}
\usepackage{amsfonts}
\usepackage{graphicx}
\usepackage{algorithm}
\usepackage[noend]{algpseudocode}
\usepackage{flushend}
\usepackage{multicol}\usepackage{multicol}
\usepackage{dblfloatfix}    

\usepackage{lipsum}
\usepackage{flexisym}

\usepackage{color}

\usepackage{graphicx}
\usepackage{caption}
\usepackage{subcaption}

\usepackage[usenames,dvipsnames]{xcolor}

\makeatletter
\def\BState{\State\hskip-\ALG@thistlm}
\makeatother

\title{Independent Component Analysis by Entropy Maximization with Kernels}
\author{Zois~Boukouvalas, Rami~Mowakeaa, Geng-Shen~Fu
        and~T\"{u}lay~Adal{\i}

\thanks{Z. Boukouvalas is with the Department
of Mathematics and Statistics, University of Maryland Baltimore County, (e-mail: zb1@umbc.edu)}
\thanks{R. Mowakeaa, G.-S. Fu, and T. Adal{\i} are with the Department 
of Computer Science and Electrical Engineering, University of Maryland Baltimore County,(e-mail: \{ramo1, fugengs1, adali\}@umbc.edu)}

}

\begin{document}

\maketitle

\begin{abstract}

Independent component analysis (ICA) is the most popular method for blind source separation (BSS) with a diverse set of applications, such as biomedical signal processing, video and image analysis, and communications. Maximum likelihood (ML), an optimal theoretical framework for ICA, requires knowledge of the true underlying probability density function (PDF) of the latent sources, which, in many applications, is unknown. ICA algorithms cast in the ML framework often deviate from its theoretical optimality properties due to poor estimation of the source PDF. Therefore, accurate estimation of source PDFs is critical in order to avoid model mismatch and poor ICA performance. In this paper, we propose a new and efficient ICA algorithm based on entropy maximization with kernels, (ICA-EMK), which uses both global and local measuring functions as constraints to dynamically estimate the PDF of the sources with reasonable complexity. In addition, the new algorithm performs optimization with respect to each of the cost function gradient directions separately, enabling parallel implementations on multi-core computers. We demonstrate the superior performance of ICA-EMK over competing ICA algorithms using simulated as well as real-world data.
\end{abstract}


\IEEEpeerreviewmaketitle

\section{Introduction}

\IEEEPARstart{I}{ndependent component analysis} (ICA) is a data-driven technique for decomposing a given set of observations into a set of statistically independent components. A natural way to achieve these decompositions is by maximum likelihood (ML) estimation which enables one to take into account all forms of diversity of the dataset described through its statistical properties. Additionally, ML theory possesses many theoretical advantages allowing the study of asymptotic optimality of the estimator, derivation of a lower bound on variance (Cram\`{e}r-Rao lower bound), and identifiability conditions \cite{6784026,Comon_HandbookBSS_2010}. Most ICA algorithms can be derived as special cases of the ML cost function \cite{6784026, 4531181}. However, knowledge of the underlying probability density function (PDF) of the latent sources is generally unknown. Algorithms that utilize a fixed model for the underlying distribution of the latent sources or a simple model---i.e., one that is not sufficiently flexible---yield poor separation performance when the data deviates from the assumed model. In such cases, these algorithms also diverge from the desirable optimality conditions of the ML estimation. The main contribution of this paper is to introduce a new flexible ICA algorithm that uses a PDF estimator that closely adheres to the underlying statistical description of the data yielding superior separation performance while maintaining the desirable optimality of ML estimation.

Among the widely used ICA algorithms, FastICA \cite{hyvarinen1999fast}, efficient fast ICA (EFICA) \cite{koldovsky2009blind}, and information maximization (Infomax) \cite{bell1995information}, use a fixed nonlinearity or model for the underlying distribution of the sources, which makes them computationally attractive but their separation performance suffers when the density of the data deviates from the assumed underlying model. Robust, accurate, direct ICA (RADICAL) \cite{learned2003ica} is a nonparametric ICA algorithm using spacings estimates of entropy. However, nonparametric methods are practically difficult due to the parameter selection that is required and are computationally demanding when number of samples increases. ICA by entropy bound minimization (ICA-EBM) \cite{li2010independent} provides flexible density matching through use of four measuring functions based on the maximum entropy principle. Four measuring functions are used for calculating the entropy bound, but the associated maximum entropy density is limited to bimodal, symmetric or skewed, heavy-tailed or not heavy-tailed distributions, which might be limited in scenarios where the PDF of the latent sources is unknown and complicated.

In this paper, we propose a new and efficient ICA algorithm, ICA by entropy bound maximization with kernels (ICA-EMK), that utilizes both global as well as {\it adaptive} local measuring functions to gain insight into the local behavior of source PDFs with a reasonable increase in model complexity. By taking advantage of the decoupling trick \cite{li2007nonorthogonal}, the optimization procedure of ICA-EMK is performed in a parallel fashion allowing the computation time to become not only a function of the number of sources, but also (inversely) the number of available processing cores.

The remainder of this paper is organized as follows. In Section II, we provide the necessary background for ICA and the principle of maximum entropy. In Section III, we provide the mathematical formulation of the proposed ICA algorithm along with its pseudocode. We also discuss its parallel implementation and demonstrate its effectiveness through a computational example. In Section IV, we demonstrate the effectiveness of ICA-EMK through simulated data as well as mixtures of face images. The conclusion is presented in Section VI. 
\section{Background} \label{background}
\subsection{Independent Component Analysis}
Let $N$ statistically independent, zero mean, and unit variance latent sources ${\bf s}(t) = [ s_1(t),\dots,s_N(t)]^{\top}$ be mixed through an unknown invertible mixing matrix ${\bf A}\in \mathbb{R}^{N\times N}$ so that we obtain the mixtures ${\bf x}(t) = [ x_1(t),\dots,x_N(t)]^{\top}$, through the linear model 
\begin{equation}\nonumber
{\bf x}(t) = {\bf A}{\bf s}(t),~~~ t=1,\dots,T,
\end{equation}
where $t$ denotes the discrete time index and $(\cdot)^{\top}$ the transpose. The goal of ICA is to estimate a demixing matrix ${\bf W}\in \mathbb{R}^{N\times N}$ to yield maximally independent source estimates ${\bf y}(t) = {\bf W}{\bf x}(t)$. This can be achieved by minimizing the mutual information (MI) of the estimated sources, which is defined as the Kullback-Leibler (KL)-distance between the joint source density and the product of the marginal estimated source densities. Thus under the assumptions that the samples are independent and identically distributed (i.i.d), the MI cost function is given by 
\begin{align}\label{ICAmutualInfo}
J({\bf W}) &= E\left\{-\log\left[\frac{p_{s_1}(y_1) p_{s_2}(y_2)\cdots p_{s_N}(y_N)}{p_{s_1s_2\dots s_N}(y_1,y_2,\dots,y_N)}\right]\right\}\nonumber \\ \nonumber
&= \sum_{n=1}^N H(y_n) - H({\bf y})\\ \nonumber
&= \sum_{n=1}^N H(y_n) - \log|\det({\bf W})| - H({\bf x}), \nonumber
\end{align}
where $H(\cdot)$ is the differential entropy and $H({\bf x})$ is a term independent of ${\bf W}$ and can be treated as a constant $C$. The MI equation is derived using that $p_{{\bf s}}\left( {\bf W}{\bf x}\right) = p_{{\bf x}}({\bf x})|\det({\bf W})|^{-1}$ and that the differential entropy of a linear transformation is $H({\bf W}{\bf x}) = \log|\det({\bf W})| + H({\bf x})$. Minimization of mutual information (MI) among the estimated sources is equivalent to the maximization of the log-likelihood cost function as long as the model PDF matches the true latent source PDF \cite{6784026}. As the model deviates from the true PDF, a bias is introduced in the estimate of the demixing matrix that can be quantified using the KL distance between the true and the estimated PDF. This can be avoided by integrating a flexible density model for each source into the ICA framework in order to minimize estimation bias of the demixing matrix yielding accurately separated sources from a wide range of PDFs.

To achieve this, the cost function and its gradient can be rewritten with respect to (w.r.t.) each ${\bf w}_m,\ m=1,\dots N$. This can be performed via the decoupling trick \cite{li2010independent}, which allows the expression of the volume of the $N$-dimensional parallelepiped spanned by the rows of ${\bf W}$ as the inner product of the $m$th row and the unit length vector vector ${\bf h}_m$ that is perpendicular to all row vectors of ${\bf W}$ except of ${\bf w}_m$. Therefore the cost function can be written as 
\begin{equation}\label{ICAmutualInfoDecoupling}
J({\bf W}) = \sum_{n=1}^NH(y_n) - \log\left| {\bf h}_m^{\top} {\bf w}_m \right| - C, ~~~m = 1,\dotsc,N.
\end{equation}
Then, the gradient of (\ref{ICAmutualInfoDecoupling}) can be written in a decoupled form and is given by 
\begin{equation}\label{gradcost}
\frac{\partial J({\bf W})}{\partial {\bf w}_m} = -E\left\{ \phi(y_m) {\bf x} \right\} - \frac{{\bf h}_m}{ {\bf h}_m^{\top} {\bf w}_m},
\end{equation}
where $\phi(y_m) = \frac{{\partial \log p(y_m)}}{{\partial y_m}}$ is called the score function. Therefore, the estimate of ${\bf W}$ can be determined w.r.t. each row vector ${\bf w}_m$, $m=1,\dotsc,N$ independently. As can be seen in (\ref{gradcost}), each gradient direction depends directly on the corresponding estimated source PDF. The sub-optimal gradient directions can lead to slower or sub-optimal convergence, or, in extreme cases, divergence of the source separation algorithm. This again justifies the desire for flexible and accurate source PDF estimates.

\subsection{Maximum Entropy Principle} \label{EMK}
Classical density estimation techniques can be characterized as either parametric or nonparametric. Parametric methods provide a simple form for the PDF and are computationally efficient, however are limited when the underlying distribution of the data deviates from the assumed parametric form. On the other hand, nonparametric methods are not limited to any specific distribution, but are computationally demanding and they highly depend on the choice of tuning parameters. In contrast, semiparametric methods, such as those based on the maximum entropy principle \cite{li2010independent,fu2015density,5946905}, combine the simple density form and the flexibility of nonparametric and parametric methods, respectively.

The maximum entropy principle, can be described by the optimization problem \cite{cover2012elements}
\begin{equation}\label{MaxEntDistOptGeneral}
\begin{aligned}
& \underset{p(x)}{\text{max}}
& & H(p(x)) = -\int_{-\infty}^{\infty} p(x) \log p(x)\ dx \\
& \text{s.t.}
& & \int_{-\infty}^{\infty} r_i(x) p(x)\ dx = \alpha_i, \text{ for } i=1,\dots, M, 
\end{aligned}
\end{equation}
where $r_i(x)$ are the measuring functions, $\alpha_i = \sum_{t=1}^T r_i(t) / T$ are the sample averages, and $M$ denotes the total number of measuring functions. To ensure that $p(x)$ is a valid PDF, we select $\alpha_1 = r_1(x) = 1$. The constrained optimization problem (\ref{MaxEntDistOptGeneral}) can be written as an unconstrained one through the Lagrangian function:
\begin{equation}\label{Lagrangian}
L(p(x)) = H(p(x)) + \sum_{i=1}^M \lambda_i \int_{-\infty}^{\infty} (r_i(x) - \alpha_i)p(x)\ dx,
\end{equation}
where $\lambda_i$, $i=1,\dots, M$ are the Lagrange multipliers. By differentiating (\ref{Lagrangian}) with respect to $p(x)$ and setting its derivative equal to zero, the equation of the maximum entropy distribution is obtained as 
\begin{equation}\label{pdfestimate}
{\hat p}(x) = \exp \left\{ -1 + \sum_{i=1}^M \lambda_ir_i(x)\right\},
\end{equation}
and the Lagrange multipliers can be numerically determined to satisfy the constraints in (\ref{MaxEntDistOptGeneral}). 
%

\section{ICA-EMK}\label{IcaEmk}
Entropy maximization with kernels (EMK) is a robust semiparametric method that has been shown to provide desirable estimation performance \cite{fu2015density}. Its flexibility to model a wide range of distributions and its simple mathematical form make it a particularly attractive candidate for ICA. EMK is able to achieve this desirable performance by using both global as well as adaptive local measuring functions to provide constraints on the overall statistics and gain insight into the local behavior of the source PDFs, respectively.

\subsection{Mathematical Formulation}
One of the main components of EMK is the numerical estimation of the Lagrange multipliers given in (\ref{pdfestimate}). For a given set of measuring functions, the Lagrange multipliers for the $m$th source estimate $y_m$, are estimated using the Newton iteration method \cite{fu2015density}
\begin{equation}\label{NewtonIter}
{\boldsymbol \lambda}^{(k+1)} = {\boldsymbol \lambda}^{(k)} - {\bf J}^{-1} E_{p^{(k)}}\left\{ {\bf r} - {\boldsymbol \alpha} \right\},
\end{equation}
where $p^{(k)}$ it the estimated PDF for the kth iteration, and ${\bf r}$, ${\boldsymbol \lambda}$, ${\boldsymbol \alpha}$ denote the $M$th dimensional vector of measuring functions, Lagrange multipliers, and sample averages respectively. The $(i,j)$th entry of the Jacobian matrix ${\bf J}$ is $\int_{-\infty}^{\infty}r_i(y_m)r_j(y_m)p^{(k)}(y_m) dy_m$ and the $i$th entry of $E_{p^{(k)}}\left\{ {\bf r} - {\boldsymbol \alpha} \right\}$ is $\int_{-\infty}^{\infty} (r_i(y_m) - \alpha_i)p^{(k)}(y_m) dy_m$. 

We select the global measuring functions $\{1, y_m, y_m^2, y_m/(1 + y_m^2)\}$ to relate to sample estimates of the PDF, mean, variance, and higher order statistics respectively. For local measuring functions, we use a number of Gaussian kernels $\{ \exp\left( -(y_m - \mu_i)^2 / 2\sigma_i^2 \right)\}$ with $i=M-4$. The number of local measuring functions is chosen by an information-theoretic criterion, the minimum description length (MDL) \cite{rissanen1978modeling,wang1998quantification}. For each Gaussian kernel the parameters $\mu$ and $\sigma^2$ are estimated by finding the greatest deviation between the estimated and the true PDF. For further details about the choice of the local measuring functions and the estimation of their parameters, we refer the reader to \cite{fu2015density}.

Our main contribution is the derivation of a new ICA algorithm, ICA-EMK that takes advantage of the accurate yet analytically simple estimation capability of EMK and yields an algorithm with superior separation performance. By using the Lagrange multiplier estimates from (\ref{NewtonIter}), the differential entropy of the $m$th source estimate can be written as
\begin{equation}\label{ICAmutualInfoEMK}
H(y_m) = -E\left\{-1 + \sum_{i=1}^M \lambda_ir_i(y_m)\right\} = 1 - \sum_{i=1}^M \lambda_i\alpha_i,\nonumber
\end{equation}
which allows us to rewrite the decoupled cost function (\ref{ICAmutualInfoDecoupling}) as 
\begin{equation}\label{ICAmutualInfoEMK}
J({\bf W}) = \sum_{n=1}^N\left(1 - \sum_{i=1}^M \lambda_i(n) \alpha_i(n)\right) - \log\left| {\bf h}_m^{\top} {\bf w}_m \right| - C,
\end{equation}
where $\lambda_i(n)$ and $\alpha_i(n)$ denote the estimated Lagrange multipliers and sample averages for each of the source estimates. The gradient of (\ref{ICAmutualInfoEMK}) w.r.t. ${\bf w}_m$ is given by 
\begin{equation}\label{gradcostdecoup}
\frac{\partial J({\bf W})}{\partial {\bf w}_m} = -\sum_{i=1}^M \lambda_i E\left\{ \frac{{\partial r_i(y_m)}}{{\partial y_m}} {\bf x} \right\} - \frac{{\bf h}_m}{ {\bf h}_m^{\top} {\bf w}_m}.
\end{equation}
Performing the optimization routine in a Riemannian manifold rather than a classical Euclidean space provides important convergence advantages. Therefore, following \cite{li2010independent}, we define the domain of our cost function to be the unit sphere in $\mathbb{R}^N$. Then, by using the projection transformation onto the tangent hyperplane of the unit sphere at the point $\mathbf{w}_m$, the normalized gradient of our cost function is given by 
\begin{equation}\label{projectionDer}
{\bf u}_m = {\bf P}_m({\bf w}_m) \frac{\partial J({\bf W})}{\partial {\bf w}_m},
\end{equation}
where ${\bf P}_m({\bf w}_m) = {\bf I} - {\bf w}_m{\bf w}_m^T$ and $||{\bf w}_m|| = 1$. A pseudo-code description of the ICA-EMK algorithm is given in Algorithm \ref{algor} below. The main part of this algorithm is the loop described in lines 3--10. Since ICA-EMK cost function depends on the number of measuring functions chosen for each source, non-monotonic behavior is expected between two consecutive iterations. The algorithm terminates when $|J(\mathbf {W}_{iter}) - J(\mathbf{W}_{iter-k})| < \delta$, where $\delta$ is a tolerance chosen by the user and $k$ is a small integer that desensitizes the algorithm to rapid changes in the cost function. The loop also terminates if the number of iterations exceeds a pre-defined maximum number of iterations.
\begin{algorithm}
\caption{ICA-EMK}\label{algor}
\begin{algorithmic}[1]
\State {\bf Input}: ${\bf X} \in \mathbb{R}^{N\times T}$
\State Initialize ${\bf W}_0 \in \mathbb{R}^{N\times N}$

	\For {$m$ = 1:$N$}
\State 	Given $\{r_i\}_{i=1}^M$, estimate Lagrange multipliers using (\ref{NewtonIter})
\State 	Compute ${\bf h}_m$, orthogonal to ${\bf w}_i$ for all $i\neq m$
\State	Calculate the derivative $\frac{\partial J({\bf W})}{\partial {\bf w}_m}$ using (\ref{gradcostdecoup})
\State	Project the gradient onto the unit sphere using (\ref{projectionDer})
\State 	$({\bf w}_m)^{\rm new} \leftarrow ({\bf w}_m)^{\rm old} - \gamma {\bf u}_m$
	\EndFor
\State {\bf end}

\State	Repeat steps 3 through 10 until convergence in $J({\bf W})$ or until the maximum number of iterations is exceeded

\State {\bf Output}: ${\bf W}$
\end{algorithmic}
\end{algorithm}

\subsection{Parallel Implementation and Performance}
In many applications encountered in practice, the number of sources can be quite large subjecting traditional sequential source separation algorithms to lengthy execution times. Since the bulk of the computational complexity of ICA-EMK occurs in lines 3--10 in Algorithm \ref{algor}, distributing separate iterations of the main loop to separate computation resources is desirable to reduce the total execution time.

The performance improvement to be gained from using a faster mode of execution is limited by the fraction of the time the faster mode can be used. This is known as Amdahl's Law and is given by \cite{hennessy}:

\begin{equation}\label{Amdahl}
\text{Speedup} = \frac{t_{\text{old}}}{t_{\text{new}}} = \frac{1}{(1-f) + \frac{f}{s}},
\end{equation}
where $t_{\text{old}}$ is the execution time prior to the enhancement, $t_{\text{new}}$ is the execution time after the enhancement, $f\leq 1$ is the fraction of $t_{\text{old}}$ spent on the code to be enhanced, and $s\geq 1$ is the speedup of the enhanced code. In repeated experimental runs of ICA-EMK, $f$ was found to be quite high, on the order of $f > 0.95$ leading to a speedup on the order of $s$.

The decoupling trick provides independence between the computation of each of the cost function gradient directions allowing for a direct exploitation of the natural parallelism on multi-processor or multi-core computers. This is performed by outsourcing the computation of each gradient direction (\ref{gradcostdecoup}) to a separate processor or core subject to availability of the computing resource. The results from the separate cores are joined in each iteration to evaluate the termination criterion. The overhead associated with forking, then joining, execution streams leads to $s$ being less than, yet very close to, $L$, the number of cores or processors available for parallel execution. Despite the noted overhead, real world applications with a sufficiently large number of sources and samples do achieve significant speedup as our experimental results in section IV demonstrate.

%

\begin{figure}[t*]
\centering
\includegraphics[width=3.2in]{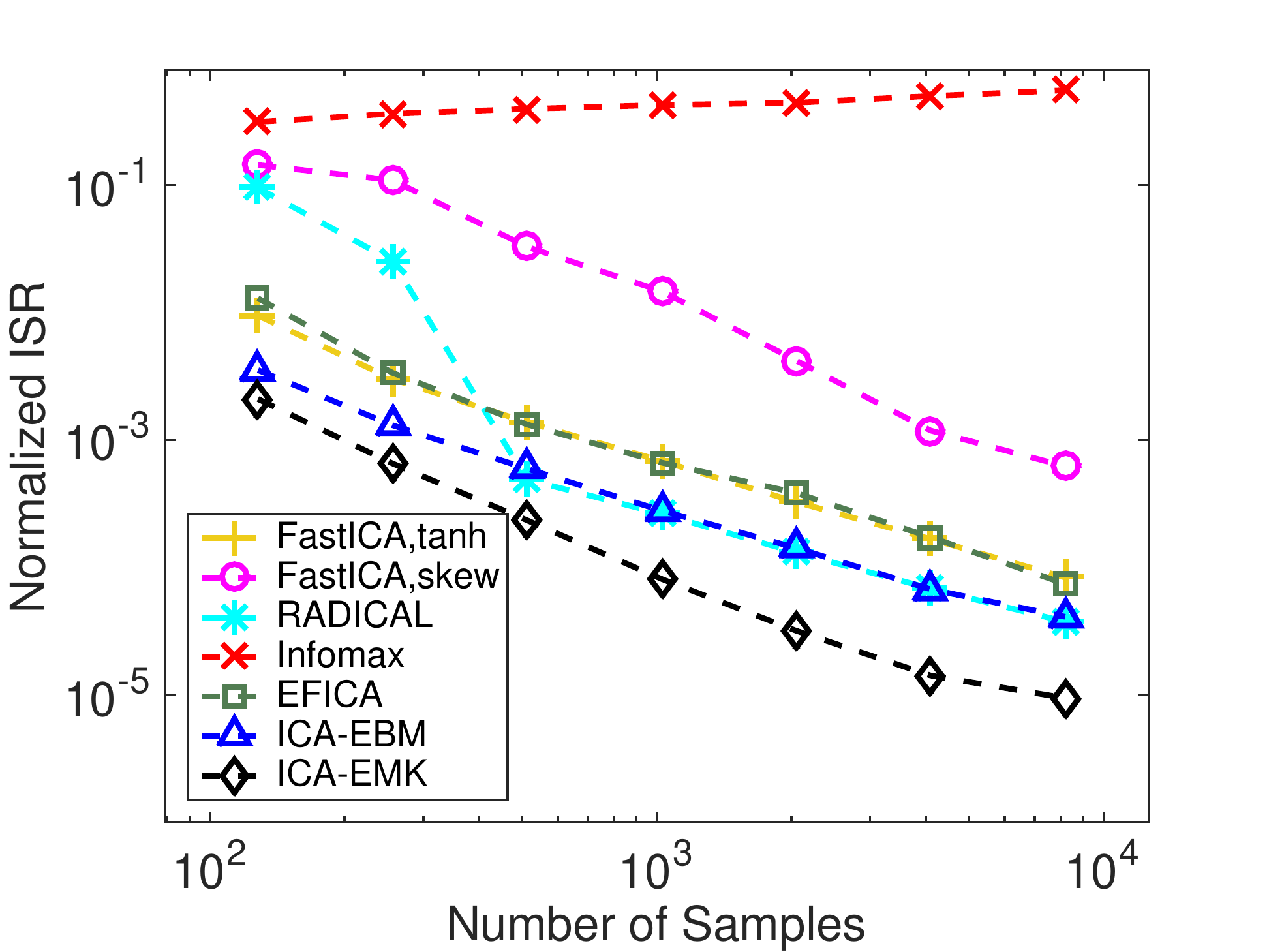}
\caption{Performance comparison of seven ICA algorithms in terms of the normalized average ISR as a function of the number of samples. The $N=8$ sources are mixtures of GGDs. Each point is the result of 100 independent runs.}
\label{fig:Mixture5}
\end{figure}
\section{Experimental Results} \label{experiments}
We demonstrate the performance of ICA-EMK, in terms of its separation power, using simulated as well as natural images as sources. We compare ICA-EMK with six commonly used ICA algorithms: 
FastICA, using the symmetric decorrelation approach with two nonlinearities tanh and skew (FastICAtanh) and (FastICAskew), RADICAL, Infomax, EFICA, and ICA-EBM. FastICAtanh favors symmetric distributions and FastICAskew skewed ones. RADICAL is a nonparametric algorithm that can successfully accommodate more complex PDFs. Infomax is based on a fixed super-Gaussian source model. EFICA is an efficient FastICA version that uses the univariate generalized Gaussian distribution (GGD) source model. ICA-EBM favors distributions that are skewed, heavy or light-tailed and bimodal. Moreover, we also quantify the performance of ICA-EMK, in terms of execution time, using simulated data. We elect to limit the maximum number of local measuring functions to 5 so as to control complexity. We observed that the overall impact of this limitation in terms of performance is negligible. In all of the following experiments, we ICA-EMK is initialized using the output of ICA-EBM.
\begin{figure}[t*]
\centering
\includegraphics[width=3.2in]{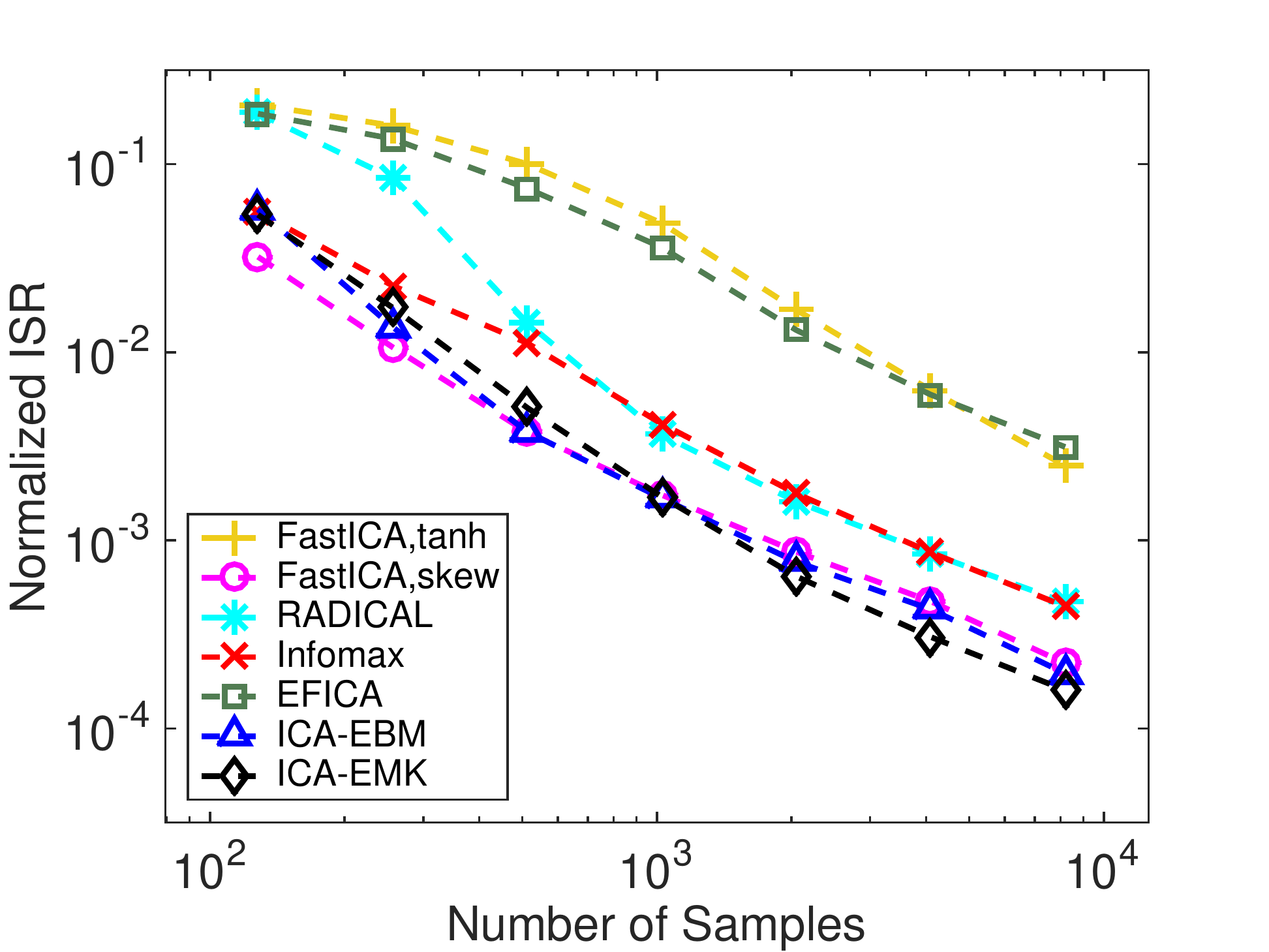}
\caption{Performance comparison of seven ICA algorithms in terms of the normalized average ISR as a function of the number of samples. The $N=8$ sources are drawn from the Gamma distribution with different shape parameters. Each point is the result of 100 independent runs.}
\label{scatter}
\label{fig:GammaSources}
\end{figure}

\begin{figure*}[t*]
\centering
  \includegraphics[width=7.3in]{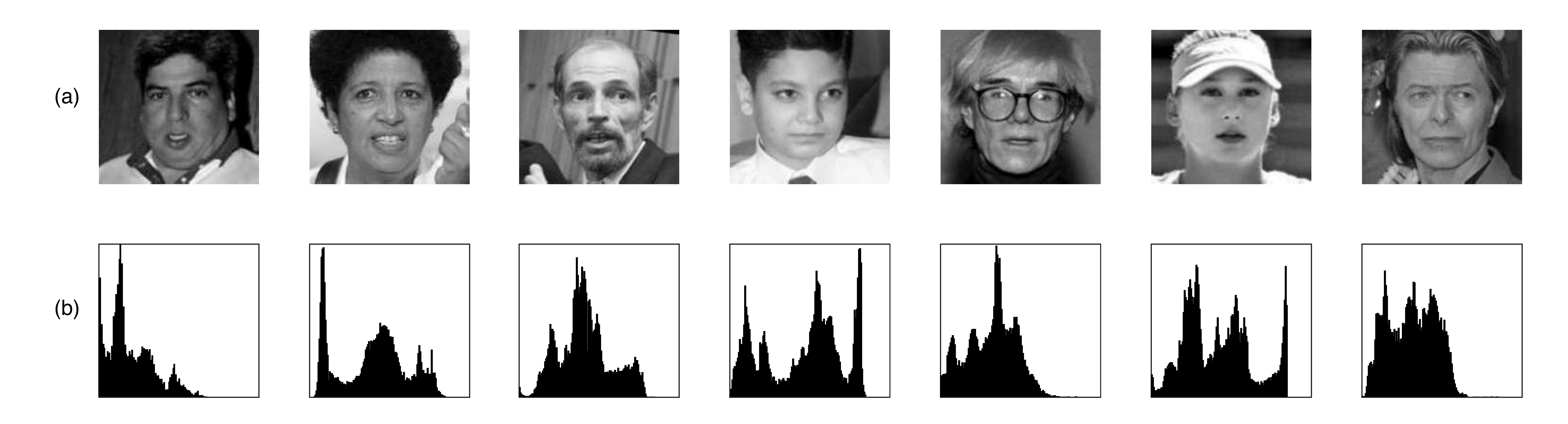}
  \caption{Seven face images with complicated densities. (a) Original grayscale sources images of size $168\times168$, (b) Histogram of each image where number of bins is 128.}
\label{fig:Histograms}
\end{figure*}

\subsection{Simulated Data}
In the first experiment, we generate 8 simulated sources each of which is a mixture of GGD kernels. The PDF of each source is given by \cite{nadarajah2005generalized}
\begin{equation}
p(x;\beta_i, \mu_i, \sigma_i) = \sum_{i=1}^K \pi_i\eta_i\exp\left( -\frac{(x-\mu_i)^{2\beta_i}}{2\sigma_i^{2\beta_i}} \right),\ x\in\mathbb{R} \nonumber
\end{equation}
where $\eta = \frac{\beta}{2^{\frac{1}{2\beta}}\Gamma(\frac{1}{2\beta})\sigma}$ and $K$ is the number of mixtures. To generate sufficiently complicated sources, $K$ is randomly selected to be either 4 or 5. The weight parameters $\pi_i$ are randomly selected from the interval $(0,1)$ such that $\sum_{i=1}^K \pi_i= 1$. The shape parameter $\beta$ is randomly selected from the interval $(0.25,4)$. Note that $\beta$ controls the peakedness and spread of the distribution. If $\beta<1$, the distribution is more peaky than Gaussian with heavier tails, and if $\beta > 1$, it is less peaky with lighter tails. When $K=4$, the GGD means are chosen to be $\{-8,-4,4,8\}$, whereas when $K=5$ the means are chosen to be $\{ -10,-5,0,5,10\}$. In the second experiment, we generate 8 sources using the Gamma distribution with PDF $p(x) = x^{\beta - 1}\exp(-x)$, $x\geq 0$. For each of the sources the shape parameter takes values from the set $\{1,2,\dots,8\}$, resulting in different unimodal skewed PDFs. For both experiments, the sources are mixed by a random square matrix whose elements are drawn from a zero mean, unit variance Gaussian distribution. To evaluate the performance of our algorithm we use the average interference to signal ratio (ISR) as in \cite{li2010independent}. The rest of the algorithm parameters are $\gamma=0.01, k = 8$, and $M = 9$. Results are averaged over 100 independent runs. 

In Fig.\ref{fig:Mixture5} we observe that ICA-EBM and RADICAL exhibit good performance as the sample size increases revealing the flexibility of their underlying density models. On the other hand, 
the two different versions of FastICA, EFICA, and Infomax do not perform well due to their simple underlying density model. Overall however, ICA-EMK performs the best among the seven algorithms.

In Fig. \ref{fig:GammaSources}, we see that FastICAskew performs the best when the sample size is less than 1000. When the sample size becomes greater than 1000, the performance of ICA-EBM is similar to that of FastICAskew since the large sample size allows for accurate approximation of the differential entropy of the estimated sources. For smaller sample sizes and simpler distributions, RADICAL does not perform well. When the sample size becomes greater than 1000, its performance is very similar to Infomax's performance. FastICAtanh and EFICA do not provide good performance compared to the other algorithms due to the inherent model mismatch. Finally, ICA-EMK for large sample sizes provides the best performance since the probability density model is most accurately estimated at each ICA iteration.

Despite its superior separation performance, ICA-EMK is computationally demanding compared with other algorithms---with the exception of RADICAL, which is the most costly for large sample sizes. The additional time penalty incurred diminishes, however, with the increase in parallelism of the computing resource at hand. The trade-off between superior performance and diminishing time penalty, hence does favor the use of ICA-EMK over others for improved performance.

\subsection{Mixture of Atificial Images}
In this experiment, ICA-EBM uses FastICA with nonlinearity function $x^4$ to provide an initial value for ${\bf W}$. Therefore, to show the improvement that ICA-EMK provides over FastICA and ICA-EBM, we use seven face images as independent sources. Fig. \ref{fig:Histograms} shows the grayscale images obtained from \cite{wolf2011effective,huang2007labeled} as well as their associated histograms. It is clear from the histograms that the images represent a wide range of complicated source distributions.

\begin{figure}[ht*]
\centering
\includegraphics[width=3.2in]{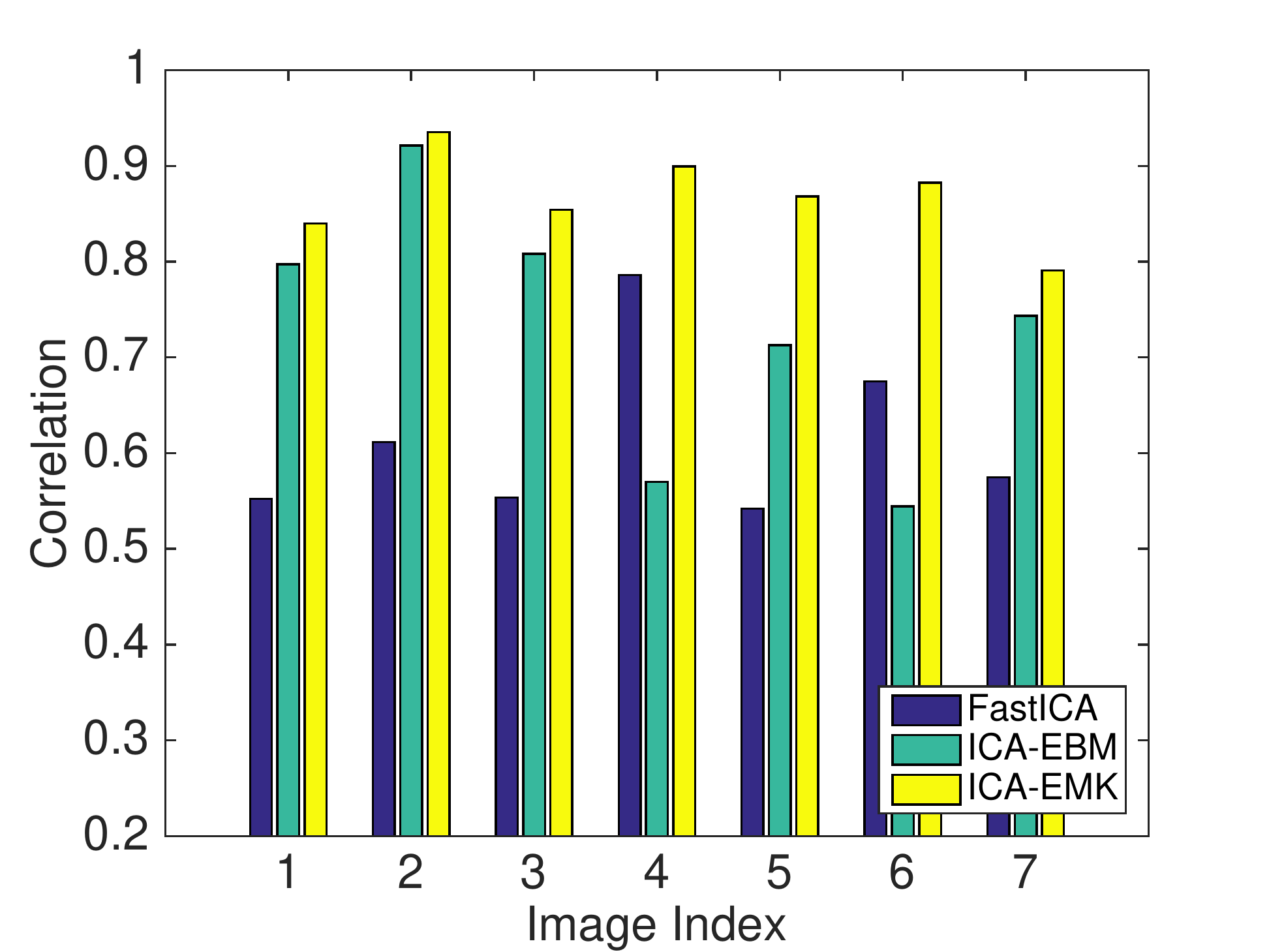}
\caption{Correlation between the true and estimated source images using FastICA (blue bars), ICA-EBM (green bars), and ICA-EMK (yellow bars) algorithms. Results are averaged over 300 independent runs.}
\label{scatter}
\label{fig:Correlation1}
\end{figure}

To setup the experiment, we create the independent sources by vectorizing the $168\times168$ images which are then linearly mixed using a random mixing matrix. After obtaining the estimated demixing matrices from each of the algorithms, we estimate the independent components and, together with their associated demixing vectors, pair them with the true sources. In the case where more than one estimated component is paired with a single true source, we use Bertsekas algorithm \cite{bertsekas1988auction} to find the best assignment as described in \cite{rodriguez2014general}. To evaluate the performance of the three algorithms, we use the absolute value of the correlation between the true and the estimated sources. Results are averaged over 300 independent runs. In Fig. \ref{fig:Correlation1}, we observe that ICA-EBM performs significantly better than FastICA for all but two images. Overall, ICA-EMK provides the best performance among the three algorithms.

\subsection{Parallel Implementation Performance}
\begin{figure}[ht*]
\centering
\includegraphics[width=3.2in]{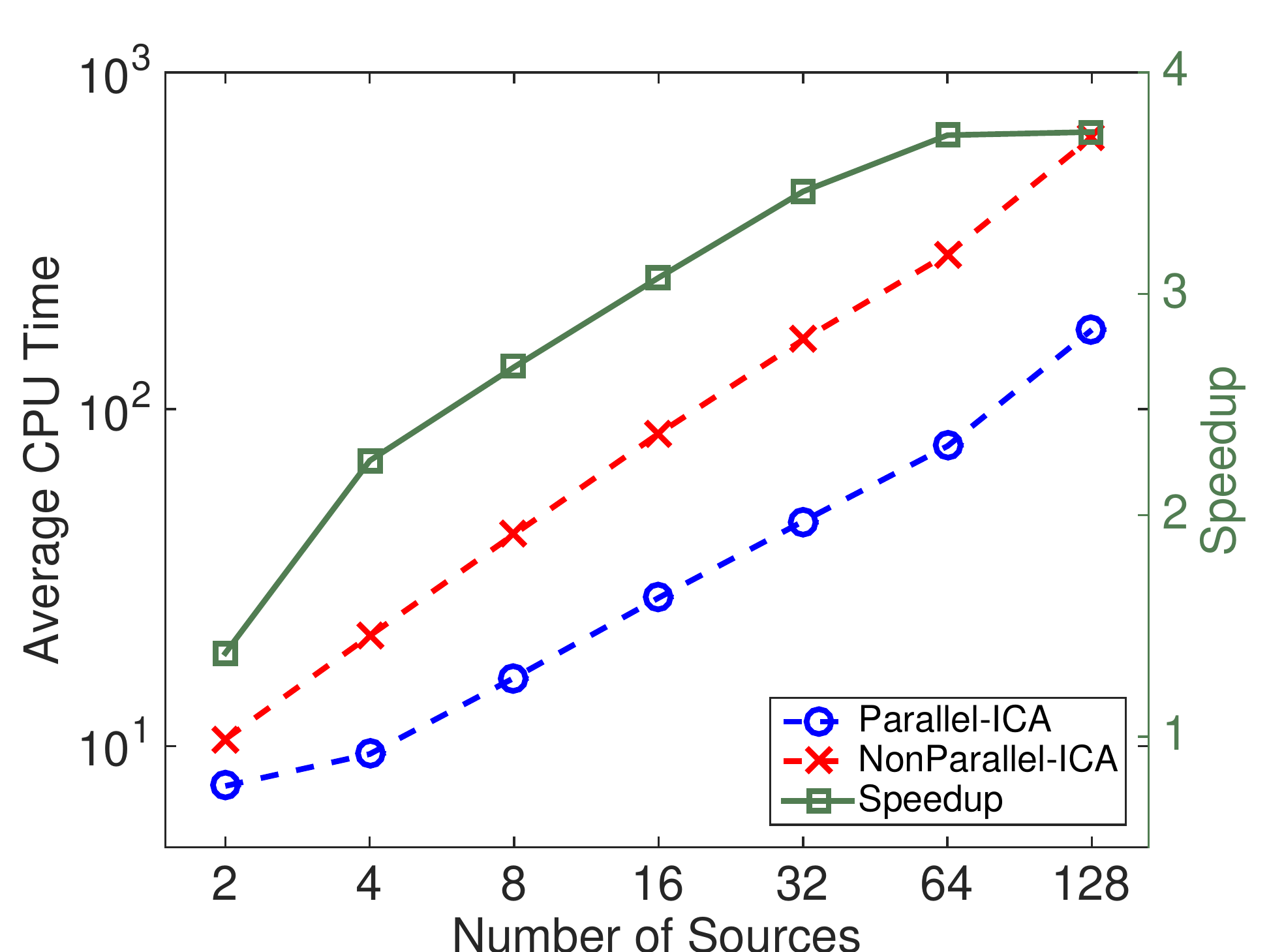}
\caption{Average CPU time for parallel and sequential implementations of ICA-EMK as a function of the number of sources and the resulting speedup.}
\label{scatter}
\label{fig:res1}
\end{figure}

To demonstrate the computational speedup of the parallel ICA implementation over its sequential counterpart (ICA-EMK where the decoupled source computations are forced to run on a single processing core), we compare the average execution time of each implementation on the 5 GGD mixtures of simulated sources from the prior subsection with $T=1000$ samples. Both implementations are performed in the Matlab environment on a lab computer with a quad-core processor and 8 GB of RAM. 
Figure \ref{fig:res1} shows the result of running this experiment for a number of sources $N \in \{2, 4, 8, 16, 32, 64, 128\}$ where each point is the result of the average of 100 independent runs. Both algorithms execute 100 iterations irrespective of convergence properties. The red and blue curves, associated with the y-axis to the left, represent the average CPU time for the non-parallel ICA and parallel ICA implementations respectively. The green curve, associated with the y-axis to the right, represents the speedup as a result of exploiting parallelism. We observe that when the number of sources $N=2$, the speedup is small since two of the four cores are idling. As the number of sources increases, the speedup improves and approaches the number of processor cores without reaching it--inline with our discussion following equation (\ref{Amdahl}). This is due to the overhead associated with forking then joining the computation in addition to the fact that the ICA algorithm is not fully parallelizable and some sequential portions remain. Similarly, with $L$ CPUs, a speedup just shy of $L$ can be expected as long as there is a sufficiently large number of sources to keep processor utilization near $100\%$. 

\section{Discussion}
In this paper, we present a new and efficient ICA algorithm, ICA by entropy maximization with kernels, that uses both global and local measuring functions to provide accurate estimates of the PDFs of the source estimates. ICA-EMK has been implemented in a parallel fashion so that it is computationally attractive when the number of sources and number of cores increase. Experimental results confirm the attractiveness of the new ICA algorithm that can separate sources from a wide range of distributions. 

Due to its flexibility, ICA-EMK can be used in many applications especially when prior knowledge about the data is not available. Where prior knowledge of the PDF exists, however, the estimation technique can be adjusted based upon the needs of the application. 

\bibliographystyle{ieeetr}
\bibliography{./IEEE_Neural_Networs_Learning_Systems_2016}

\end{document}